\documentclass[a4paper]{article}
\usepackage[T1]{fontenc}
\usepackage{pgf}
\usepackage{url}
\usepackage[english]{babel}
\usepackage{multirow}
\usepackage{fancyhdr}
\usepackage{anysize}
\usepackage{amssymb}
\usepackage{graphicx}
\usepackage{url}
\usepackage{algorithm}
\usepackage{algpseudocode}
\usepackage[sort, numbers]{natbib}
\usepackage{ntheorem}
\usepackage{amsmath}

\begin{document}

\title{\bf{Eigenspace Method for Spatiotemporal Hotspot Detection}}
\author{Hadi Fanaee-T and João Gama}
\date{}
\maketitle
\begin{center}
Laboratory of Artificial Intelligence and Decision Support (LIAAD), University of Porto\\
INESC TEC, Rua Dr. Roberto Frias, Porto, Portugal\\
hadi.fanaee@fe.up.pt and jgama@fep.up.pt\\
\end{center}

\begin{abstract}
Hotspot detection aims at identifying subgroups in the observations that are unexpected, with respect to the some baseline information. For instance, in disease surveillance, the purpose is to detect sub-regions in spatiotemporal space, where the count of reported diseases (e.g. Cancer) is higher than expected, with respect to the population. The state-of-the-art method for this kind of problem is the Space-Time Scan Statistics (STScan), which exhaustively search the whole space through a sliding window looking for significant spatiotemporal clusters. STScan makes some restrictive assumptions about the distribution of data, the shape of the hotspots and the quality of data, which can be unrealistic for some nontraditional data sources. A novel methodology called EigenSpot is proposed where instead of an exhaustive search over the space, tracks the changes in a space-time correlation structure. Not only does the new approach presents much more computational efficiency, but also makes no assumption about the data distribution, hotspot shape or the data quality. The principal idea is that with the joint combination of abnormal elements in the principal spatial and the temporal singular vectors, the location of hotspots in the spatiotemporal space can be approximated. A comprehensive experimental evaluation, both on simulated and real data sets reveals the effectiveness of the proposed method. 
\end{abstract}

{\bf Keywords:} Hotspot Detection, Spatiotemporal Data, Eigenspace, SVD, Outbreak Detection

\section{Introduction}\label{sec:introduction}

Eigenspace techniques are very popular ones, which encompass many applications in data mining, signal processing, information retrieval and other domains. A famous instance of such an application relates to the current success of Google search engine. As described in an article entitled \textit{\$25,000,000,000 Eigenvector }\cite{bryan200625} Google search engine is largely attributed to the eigenspace techniques. Another success story is related to the application of the Singular Value Decomposition (SVD) \cite{klema1980singular} in collaborative filtering. In the year 2008, the BellKor team won the \$1,000,000 prize for the improvement of the system of the Netflix movie recommendation. The report later \cite{bell2007bellkor} stated that SVD was the key data analysis tools that they used. However, the application of the eigenspace techniques is not restricted to the above instances. There are several different examples in other areas and sciences. For instance, in face recognition \cite{turk1991eigenfaces} the specific set of the largest eigenvectors can be used to approximate the images of the human face. In structural engineering, both the eigenvalue and eigenvectors are used to estimate the vibration of structures. In control engineering, the eigenvalues of the linear system are used to assess the stability and response of the system \cite{wiki:ControlSystems}.

Nevertheless, despite the merits of the eigenspace techniques, they have not been applied yet to some potential problems, such as hotspot detection. Hotspot detection which may come with different terminologies, such as outbreak detection, cluster detection or event detection is somehow related to the clustering and anomaly detection, however it is distinct from these two. In clustering, the entire data set is partitioned into some groups, but in the anomaly detection, the anomalous points are searched for and sought after. Hotspot detection addresses the same problem, but with this difference that anomalous instances are recognized given some \textit{baseline} information. In other words, looking into the dataset, everything might seem normal, however, when the cases along the baseline are considered, some points might be considered unexpected. A realistic scenario of the application of the hotspot detection is in disease surveillance. Suppose that we have the population of different postal codesduring a range of yearsas the baseline information and the count of the reported diseases in a range of postal codes, throughout different years as the cases dataset. The goal is to detect those spatiotemporal regions that contain unexpected counts. For instance, the output like zones \textit{S1},\textit{S2} and \textit{S3} during the years \textit{T1} to \textit{T5} might be considered a spatiotemporal hotspot. The detection of such hotspots enables the officials to better understand their target of interest for essential medical care and preventive measures.   

The current methods for hotspot detection are twofold: clustering-based techniques and the scan statistics-based ones. Clustering-based techniques, such as \cite{levine2006crime} infer some thresholds from the population data and then apply the thresholds for clustering of data points in the cases set. Their prominent benefit, as opposed to the other methods, is that they provide the exact shape of the clusters. However, handling complex data, such as spatiotemporal data is not straightforward for these techniques. Besides, the clustering methods do not consider chance and randomness issues, which are very important in sensitive applications, such as security and public health. Moreover, there is not a standard clustering method for hotspot detection, which is widely accepted by the community. The methods are mostly outspread and diverse, in terms of the technical details. The clustering methods also require restrictive input parameters, which make their usage limited in automatic settings.

The second group of techniques which relies on scan statistics are widely used and accepted by the epidemiological community. These groups of techniques exhaustively scan the whole space to find interesting spatial and spatiotemporal clusters. A specific statistic is computed for each possible window and then potential clusters are sorted, based on the obtained statistics. Thereafter, the statistical significance of the top-k clusters is simulated via a Monte-Carlo simulation. Since these methods scan an entire space, they are extremely computationally expensive. Spatial scan statistics \cite{kulldorff1997spatial} requires computation time of $O(N^3)$ and space-time scan statistics (STScan) \cite{kulldorff1997spatial,kulldorff1995spatial,kulldorff1999spatial,kulldorff1998evaluating} requires $O(N^4)$. Some recent efforts are made to reduce this complexity. For instance, \cite{agarwal2006spatial} propose a method that requires $O(\frac{1}{\epsilon}N^2Log^2N)$ for spatial scan, which is more efficient than $O(N^3)$. However, the minimum complexity for space-time scan statistics in the best condition has not reached less than $O(N^3)$. This high computational cost practically has restricted their use in real-time applications or large-scale data sets. Besides, scan statistics-based techniques are highly associated with the strong parametric model assumptions (e.g. Poisson or Gaussian counts) \cite{Neill2006}. These assumptions mitigate the performance when the models are incorrect for nontraditional data sources. Additionally, scan statistics-based methods are not efficient for detection of irregular shape clusters \cite{tango2005flexibly, duczmal2004simulated} apart from the circles (spatial scan) and cylinders (space-time scan). They also assume that data is presented in a high quality format, hence is vulnerable against the noises and outliers \cite{neill2007robust}.

Our proposed method is a solution to some of the above mentioned issues, in scan statistics-based methods. An efficient method is proposed (linear with both space and time dimensions) for approximation of hotspots in the spatiotemporal space, without the need for exhaustive search. Instead of looking for deviations in the assumed parametric model, we track changes in the space-time correlation structure, using the eigenspace techniques. This approach enables us to detect irregular shape hotspots from even noisy data sets, without any prior knowledge about the data nature or hotspot characteristics. To the best of our knowledge this problem has not already been addressed by other researchers. Our approach also differs from those of ones that focus on the improvement of scan statistics-based methods efficiency (e.g. \cite{agarwal2006spatial, Neill2004, Neill2010}). We do not improve the efficiency of scan statistics-based methods; rather we propose and examine a new methodology, which follows a different aim. Hence, this is not an ``apples to apples'' comparison, as both groups of approaches have inherent differences and subsequently their own applications. STScan can be more helpful for retrospective and sensitive applications, when some prior knowledge exists about the nature of hotspot and data. On the other hand, our presented approach focuses more on real-time applications, where neither the nature of data nor the hotspot characteristics is known in advance. In such circumstances, a computationally feasible approximation method that rapidly identifies the alarming areas, without any prior knowledge might be very useful. 

The rest of the paper is organized as follows: The section \ref{sec:approachproposed} describes the problem, the proposed solution and algorithm, as well as an illustrative example. The section \ref{sec:experimentalevaluation} includes an experimental evaluation and results of the simulation study and the real case study. The last section concludes the exposition presenting the final remarks.

\section{The proposed approach} \label{sec:approachproposed}

\subsection{The problem} \label{sec:problem}

Given a spatiotemporal count matrix for the cases needed for the detection of those spatiotemporal regions (hotspots) that seem unexpected, given the baseline spatiotemporal matrix. Each cell in each matrix represents a count corresponding to a specific region and time. In particular, for disease outbreak detection, each cell in the baseline matrix represents the population corresponding to a region in a specific time period. Each cell in the matrix cases also represents the count of reported disease in a specific region, within a given time period, as well. The purpose is to determine those subgroups of the spatiotemporal space whose reported cases are unexpected. 

A baseline method that can be applied to the problem is to compute the ratio of the cases to the population for all possible spatiotemporal regions (each cell in the spatiotemporal matrix) and then compute the z-score of the ratios. Then the null hypothesis $H_o$: \textit{there is no hotspot} is rejected, in case some spatiotemporal regions with z-score greater than a threshold are found. This approach theoretically and practically  as will be illustrated later imposes too many false alarms, since for a $n \times m$ matrix, its required to perform $n \times m$ comparison tasks. In this paper, an approach that performs only $n+m$ comparisons is clearly proposed. An unsupervised method may be needed to use with the clustering on the ratios. However, it suffers from the same problem of the baseline method (it requires $n \times m$ comparisons and not $n+m$). Besides, the requirement for determination of appropriate cut-point or number of clusters adds more complexity and user involvement to the system. We are interested in developing a system, which has the following characteristics: 1) does not require any input parameter; 2) weighs all the possible hotspots, based on a standard metric like statistical significance (p-value). The benefit is this that the output can be compared to relevant systems or methods. The alpha threshold is also easy to estimate (usually alpha=0.01 or 0.05). 

\subsection{The Method}\label{sec:method}

In this section, the logic used in the method deployment is thoroughly described. Assume that we have two identical $n \times m$ matrices B (baseline) and C (cases) such that \textit{n} be the number of components in the spatial dimension and \textit{m} be the number of components in the temporal dimension. The SVD of the $n \times m$ matrix is a factorization of the form $\mathbf{M} = \mathbf{U} \boldsymbol{\Sigma} \mathbf{V}^*$. The \textit{n} columns of \textit{U} and the \textit{m} columns of \textit{V} are called the left-singular vectors and right-singular vectors of the matrices, respectively. The left-singular vectors correspond to spatial dimension, while the right-singular ones correspond to the temporal dimension. In order to clarify and elaborate, a new terminology \textit{spatial singular vector} is used along with \textsl{temporal singular vector} that respectively refers to the principal left singular vector and the principal right singular one. Note that we take only the singular vector corresponding to the largest eigenvalue for the comparison, due to the fact that the first principal singular vector represents the largest possible variance. Hence, it explains or extracts the largest part of the inertia of the data \cite{abdi2010principal}.

Now, let us denote the spatial singular vector of baseline (B) and cases (C) respectively with $\vec{\mathbf{sb}}(sb_1,sb_2,...sb_n)$ and $\vec{\mathbf{sc}}(sc_1, sc_2, ... sc_n)$. Then lets denote temporal singular vector of \textit{B} and \textit{C} respectively with $\vec{\mathbf{tb}}(tb_1, tb_2, ...tb_m)$ and $\vec{\mathbf{tc}}(tc_1, tc_2 ... tc_m)$. If we hypothetically assume that B=C, then $\vec{\mathbf{sb}}=\vec{\mathbf{sc}}$ and $\vec{\mathbf{tb}}=\vec{\mathbf{tc}}$. In this condition, the angles between $\vec{\mathbf{sb}}$ and $\vec{\mathbf{sc}}$, and between $\vec{\mathbf{tb}}$ and $\vec{\mathbf{tc}}$ would be equal to almost zero. Now assume that some change occurs in the \textit{C} and this change corresponds to a specific region and time period. Therefore, the matrices are no longer identical and subsequently the angles between their singular vectors rise up in value. From this angle change, we can only infer that some changes occur, but we do not know what subgroup of data is affected by this change. If we could identify those vector elements from $\vec{\mathbf{sc}}$ and $\vec{\mathbf{tc}}$ that caused this change, we would be able to identify the spatial and temporal components of the affected area. For instance, assume that through a hypothetical method we could identify that $sc_1$ from the $\vec{\mathbf{sc}}$ and $tc_1$ from $\vec{\mathbf{tc}}$ corresponding to the affected area. If we remove the region corresponding to $sc_1$ and time related to $tc_1$ from both baseline and cases data sets, the matrices should again become identical. Hence, we would have the angles between the pair singular vectors equal to almost zero. Here, $(sc_1, tc_1)$ is called \textit{hotspot} and $sc_1$ and $tc_1$ are respectively called the spatial and temporal components of the hotspot. The process of finding these components is also called \textit{hotspot detection}. Note that in this work, the angle between the singular vectors is not computed. In the above, the angle concept is only used to explain the rationale behind the proposed method. 

Some assumptions were made above, which were only for simplification of explanation. In practice, we rarely find two identical baseline and cases matrix. However, we are able to assume that in a normal condition, where no hotspot exists, both baseline and cases set should have a same space-time correlation structure. In this case, the pair singular vectors of baseline and cases sets, regardless of the data distribution should stay in a constant distance. Now, if a hotspot starts to grow in the cases set, this change can be directly observed from the changes in the singular vectors elements. In such cases, some distances between the singular vector elements become abnormal for elements corresponding to the affected areas in both the spatial and temporal dimension. We exploit this idea to develop our algorithm for hotspot detection. 

According to the explanation above, two kinds of tools are required, a tool for obtaining the singular vectors of a non-square matrix and a process control tool for monitoring distances between singular vector elements. SVD and statistical process control (SPC) are two powerful techniques, which for several years have been successfully applied to many problems in different domains and are state-of-the art methods to these requirements. In the next section, we explain how these techniques are exploited in the deployment of the solution. 

\subsection{The EigenSpot Algorithm}\label{sec:algorithm}

\begin{algorithm}
\caption{EigenSpot} \label{algorithm:EigenSpot}
\begin{algorithmic}[1]
\Statex //n: number of items in the spatial dimension
\Statex //m: number of items in the temporal dimension
\Statex //B: Baseline $n \times m$ spatiotemporal matrix  
\Statex //C: Cases $n \times m$ spatiotemporal Matrix
\Statex //$\alpha$: Statistical significance level (e.g. 0.05)
\Require  B, C, $\alpha$
\Ensure Hotspots
\State [$\vec{\mathbf{sb}}$,$\vec{\mathbf{tb}}$] = 1-rank SVD (B) \Comment{baseline : $\vec{\mathbf{sb}}$: spatial singular vector, $\vec{\mathbf{tb}}$:  temporal singular vector}  \label{line:svd1}
\State [$\vec{\mathbf{sc}}$,$\vec{\mathbf{tc}}$] = 1-rank SVD (C) \Comment{cases: $\vec{\mathbf{sc}}$: spatial singular vector, $\vec{\mathbf{tc}}$: temporal singular vector}  \label{line:svd2}
\For{i=1:n} \label{line:substractstart}
	\State $ds_i=sc_i-sb_i$ \Comment{ds: subtract vector corresponding spatial dimension}
\EndFor 
\For{j=1:m} 
	\State $dt_j=tc_j-tb_j$ \Comment{dt: subtract vector corresponding temporal dimension}
\EndFor \label{line:substractend}
\State $Spatial\ out\ of\ control\ elements\ \gets ControlChart (ds,\alpha)$
\State $Temporal\ out\ of\ control\ elements\ \gets ControlChart (dt,\alpha) $
\State $Hotspots \gets All\ joint\ combination\ of\ out\ of\ control\ elements\ in\ spatial\ and\ temporal\ dimensions$
\end{algorithmic}
\end{algorithm}

In this section, our proposed algorithm EigenSpot is explained in detail. The inputs of the algorithm {\ref{algorithm:EigenSpot}} are $n \times m$ matrices for the baseline and cases where \textit{n} represents the number of regions and \textit{m} represents the number of temporal instants. We start by decomposing both matrices, using one-rank SVD. The one-rank SVD gives us the principal singular vector corresponding to the spatial and temporal dimensions (lines \ref{line:svd1}-\ref{line:svd2}). The reason why the low-rank SVD is applied versus the full-rank SVD is that our approach requires only the principal singular vector for each matrix. The full-rank SVD is a more expensive method, because of the fact that a $N \times N$ matrix requires $O(N^3)$ while the low-rank SVD requires $O(kN^2)$ where in our case k=1 and therefore we require only $O(N^2)$ for each one-rank SVD. The principal singular vector explains the majority of variance in both cases and baseline. Therefore, it is appropriate for matching purposes. 

In the next step, we subtract each element of the pair singular vectors together (lines  \ref{line:substractstart}-\ref{line:substractend}). If we denote the spatial singular vector for baseline with $[sb_1\ sb_2\ ... sb_n]$ and spatial singular vector for cases with $[sc_1\ sc_2\ ... sc_n]$, the subtract vector would be $ds=[ds_1=sc_1-sb_1\  ds_2=sc_2-sb_2\  ... \ ds_n=sc_n-sb_n]$. Similarly for the temporal dimension, we have $dt=[dt_1=tc_1-tb_1\ dt_2=tc_2-tb_2\ ... dt_m=tc_m-tb_m]$. Subsequently, in order to identify the spatial and temporal components of the hotspot, a z-score control chart is applied on vectors $ds$ and $dt$ with significant level $\alpha$. To do so, the standardized vector of z-scores is first computed for $ds$ and $dt$. Thereafter, we obtain the equivalent two-tailed p-value for each z-score. Finally, those components of $ds$ and $dt$ that obtain p-value lower than $\alpha$ are considered abnormal. Finally, a joint combination of all spatial and temporal components to the original space gives us the approximation of hotspots.

For instance, assume that $\overrightarrow{sb}=[0.25\ 0.10\ 0.75\ 0.20]$ be the spatial singular vector of baseline and $\overrightarrow{sc}=[0.30\ 0.90\ 0.80\ 0.15]$ be the spatial singular vector of cases. Each element in the spatial singular vector corresponds to a specific region. For instance, 0.30 and 0.25 in the first element corresponds to region 1. Similarly, the second, third and the fourth element corresponds to the region 2, 3 and 4, respectively. The angle between the two singular vectors \textit{sb} and \textit{sc} is equal to 68 degrees in this example. This angle does not tell us what elements of singular vector have contributed to this difference. However, if in the above example, if we remove region 2 from the system, we have two vectors $\overrightarrow{sb}=[0.25\ 0.75\ 0.20]$ and $\overrightarrow{sc}=[0.30\ 0.80 \ 0.15]$ where the angle between them is equal to 0.09 which is almost equal to zero. Region 2 in this example is equivalent to the spatial component of the hotspot. In order to identify the region 2 in this example, a z-score control chart is applied on the subtract vector $\overrightarrow{ds}=[0.25-0.30\ 0.10-0.90\ 0.75-0.80\ 0.20-0.15]=[-0.05\ -0.80\ -0.05\ 0.05]$. Afterwards, we compute the standardized z-scores of the subtract vector, which in this case is $zds=[0.4119\ -1.4893\	0.4119\	0.6654]$. As shown, z-score of -1.4893 is equivalent to the left-tailed P-value of 0.06. If we define $\alpha=0.10$, region 2 would be identified as hotspot spatial component. This is because its p-value is lower than 0.10. However, if we define $\alpha=0.05$, region 2 is not detected as hotspot spatial component.

\subsubsection{The Algorithm Complexity}\label{sec:complexity}

If we assume that we have N regions and N time instants, EigenSpot requires two $O(N^2)$ for two one-rank SVD for cases and baseline matrices and two $O(N)$ for elements matching corresponding spatial and temporal dimensions. This makes the EigenSpot require only $O(2N^2)+ O(2N) = O(N^2)$ which is much more efficient than the STScan. Because, the STScan requires $O(NlogN)$ and $O(N^2logN)$ for finding the relevant time and space cylinders and $O(N^4)$ for finding the space-time cylinders as intersections of space and time cylinders \cite{assunccao2003early}. Therefore, a single execution of the STScan procedure takes $O(NlogN) + O(N^2logN) + O(N^4) = O(N^4)$. 

\subsection{Illustrative Example} \label{sec:example}

\begin{figure}
 \begin{center}
  \includegraphics[width=\textwidth]{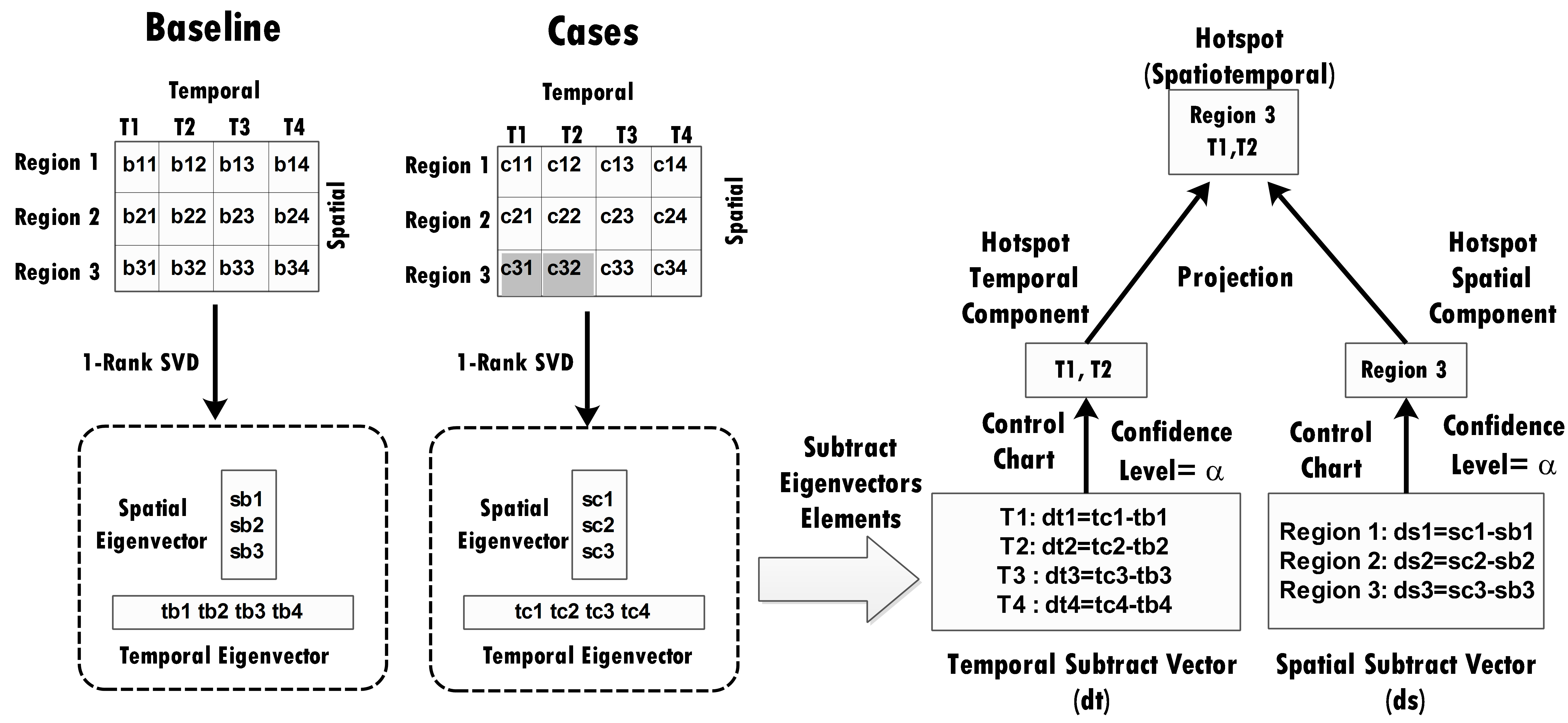}
 \end{center}
 \caption{EigenSpot Algorithm is an Illustrative example. The goal of the approach is the identification of the shaded area in the cases matrix. The values c and b in the baseline and cases matrix are counts corresponding to a spatiotemporal window. The process is composed of the following four steps: 1) matrix decomposition; 2) subtraction of pair singular vectors elements; 3) applying the z-score control chart on the subtract vector; and 4) combining the spatial and temporal hotspots components.} \label{fig:method}
\end{figure}

Figure \ref{fig:method} demonstrates an illustrative example of how a hotspot can be identified by the EigenSpot algorithm. We are given two sets of baseline and cases that encompass three regions within four time windows. Each region can be a postal code or a city. In addition, each temporal window can be a time period, such as a year (e.g. T1= 2010). If we represent these two sets as a matrix, we have two sets of $3 \times 4$ matrices such that each cell represents the count. For instance, \textit{b11} represents the population of region 1 at time window \textit{T1} and \textit{c32} represents the count of reported disease in region 3 within the temporal window \textit{T2}. The shaded area in the cases matrix (conjunction of third row with first-second columns) is the hotspot of interest that is required to be detected by the method. As demonstrated, the principal singular vector corresponding to the spatial and temporal dimensions is obtained via one-rank SVD. As a result, we have two singular vectors corresponding to the spatial and temporal dimensions for each set. In the next step, we subtract elements of each singular vectors pairs together. Therefore, we would have two vectors \textit{dt} and \textit{ds} which represent subtract vectors for the temporal and spatial dimensions, respectively. As demonstrated, \textit{dt} has four elements and \textit{ds} has three elements, each of which corresponds to the original regions and temporal windows (e.g. \textit{dt1} corresponds to \textit{T1} and \textit{ds1} corresponds to region 1). In the next step, we apply a z-score control chart with significance level $\alpha$ (e.g. $\alpha=0.05$) on both of these vectors to identify their abnormal elements. As it is hypothetically shown in the example, T1 and T2 are identified as temporal hotspot components, and region 3 is identified as the spatial hotspot component. We only need to combine spatial components with temporal components to approximate the hotspots in the spatiotemporal space. As shown, the identified hotspot of \textit{Region 3 , T1,T2} is equivalent to the shaded area in cases matrix (the target).

\section{Experimental Evaluation} \label{sec:experimentalevaluation}

In this section, the effectiveness of our proposed approach is assessed, through a comprehensive experimental study. The data sets used in evaluation of hotspot detection techniques are usually threefold \cite{BuckeridgeBCHM2005}: 1) wholly simulated: both baseline and cases and hotspots are simulated; 2) semi-realistic: baseline is taken from a real population, but cases and hotspots are simulated; and 3) real data: both baseline and cases are real and hotspots are verified by a domain specialist. In this paper, we evaluate the proposal, using the latter two strategies. We evaluate the algorithm performance, via the simulation study (section \ref{sec:simulation}) and a real-world data (section \ref{sec:real}). All experiments are conducted on a PC with Intel Core 2 Duo CPU and 3GB Ram. We use MATLAB 7 for the algorithm implementation and experiments and SatSan 9.2 \cite{satscanKulldorff2012} for experimenting with STScan. 

Our method is compared with two other techniques, including the STScan and a baseline method. STScan \cite{kulldorff1997spatial,kulldorff1995spatial,kulldorff1999spatial,kulldorff1998evaluating} exhaustively moves a varying radius and height cylinder over the whole spatiotemporal space. The height of the cylinder represents the time dimension and the surface corresponds to the space dimension. Furthermore, it scores each possible cylinder, based on likelihood ratio statistics. Next, it sorts cylinders based on an order of the highest to the lowest score. Finally, a randomization test is performed for obtaining the cylinders statistical significance. The cylinder whose p-value is lower than $\alpha$ (e.g. 0.05) is returned as hotspots. In the baseline method, we compute the ratio of the count of cases to the corresponding population for each matrix cell then we compute the z-score of the obtained ratios and obtain the p-value from z-score. Afterwards, we signal an alarm, when p-value for a cell goes lower than $\alpha$.

\subsection{Simulation Study} \label{sec:simulation}
Here, we describe how the simulated data is generated and subsequently present the obtained result. 

\subsubsection{Data Generation} \label{sec:datageneration}

We generate 1500 sets of semi-real data, based on the extracted baseline data set from \cite{kulldorff1998evaluating}. The baseline set includes the spatiotemporal distribution of population in New Mexico, USA during 1973-1991. In order to simulate the cases count, we initially obtain the maximum likelihood of the parameter of the Poisson distribution, $\lambda$ from the first year of the baseline set. Let the vector of counts for the first year be $(c_1,c_2,..,c_i)$ where $\leq<i \leq n$ (n:number of spatial items). $\lambda$ simply can be obtained by computing the means of the vector. Then, we multiply $\lambda$ by a fixed constant of 1.2\% for subsequent years (1.2\% is the average population growth rate). Next, we generate random numbers from the Poisson distribution with corresponding estimated parameters for each year. In order to inject the hotspot into the cases, we select a matrix window with size $H \times H$ (hotspot size) and multiply the counts inside the window by a fixed value of $I$ (hotspot impact).  We then vary H from 1 to 5 and select I from (1.5, 2, 2.5). Since we generate data sets based on the random numbers, we generate 100 datasets for each setting to reduce the effect of randomness. Next section explores the evaluation results.

\subsubsection{Performance Evaluation} \label{sec:evaluation}

\begin{table}
\begin{center}
\caption{The mean accuracy for 173 $\alpha$ in the range of 0.20 to 0.01 averaged for 100 data sets} \label{tab:results}

  \begin{tabular}{lllllll}
   \hline
     &  & \multicolumn{5}{c}{Size} \\ 
    Method & Impact & $1 \times 1$    &   $2 \times 2$    &   $3 \times 3$    &   $4 \times 4$    &   $5 \times 5$   \\ \hline
    \textbf{EigenSpot}  &  1.5      & 0.7011  &  0.7670 &  \textbf{0.8124}  &  \textbf{0.8574}  &  \textbf{0.8263} \\ 
    Baseline   &  1.5      & 0.7270  &  0.7417  &  0.7523  &  0.7663  &  0.7669 \\ 
    STScan    &  1.5      & \textbf{0.7966}  &  \textbf{0.7984}  &  0.8008  &  0.8030  &  0.8005 \\ \hline
    \textbf{EigenSpot}  &  2.0      & \textbf{0.8751}  &  \textbf{0.9588}  &  \textbf{0.9588}  &  \textbf{0.9492} &  \textbf{0.9498} \\ 
    Baseline   &  2.0      & 0.7259  &  0.7453  &  0.7510  &  0.7662  &  0.7741 \\  
    STScan    &  2.0      & 0.8034  &  0.8130  &  0.8171  &  0.8273 &  0.8267 \\ \hline
    \textbf{EigenSpot}  &  2.5      & \textbf{0.9393}  & \textbf{0.9718}  &  \textbf{0.9725}  &  \textbf{0.9675}  &  \textbf{0.9555} \\ 
    Baseline   &  2.5      & 0.7321  &  0.7511  &  0.7588  &  0.7783  &  0.7879 \\
    STScan    &  2.5      & 0.8069  &  0.8314  &  0.8578  &  0.8629  &  0.8723 \\

  \hline
  \end{tabular}
\end{center}
\end{table}

Hotspot detection can be considered a binary classification problem, because the detection approach marks each spatiotemporal window with hotspot or non-hotspot. However, in any approach, we determine a decision threshold to distinct hotspots from non-hotspots. Determination of this threshold becomes more important in sensitive applications, such as security and public health. For such applications, the evaluation of methods has to be evaluated within different ranges of decision thresholds. ROC curve \cite{zweig1993receiver} is a widely accepted method for such evaluation tasks. However, due to two reasons, ROC curve cannot be used as an appropriate strategy for the evaluation in this simulation study. On one hand, we want to evaluate the method performance on 100 random data sets for each 15 setting. Therefore, we have 1500 data sets, which require the analysis of 1500 ROC curves, which is infeasible. We also cannot reduce the number of data sets to one, because we are generating random sets and if we rely only on one data set then our results would be highly dependent on the chance and randomness. At the first glance, the Area Under ROC Curve (AUC) seems to be an appropriate choice, as the AUC does not have user-defined parameters. Besides, it is a summarized scalar and seems to be appropriate for mass comparison of the methods. However, the main criticism about use of AUC in applications, such as hotspot detection and outbreak detection is that AUC considers all thresholds equal, which is not true in many applications. In practice, in sensitive applications, such as epidemiology where we deal with the human lives, we are not interested in knowing how a method performs in high alpha values. The operational and practical p-value used is always low values. In other words, the alpha of interest is not between 0 and 1, rather is limited to lower values. Besides, AUC as a summarized scalar hides the real ROC curve behind the evaluation. In fact, AUC can give potentially misleading results if ROC curves cross \cite{hand2009measuring}. Some detailed criticisms against AUC can be found in \cite{hand2009measuring,hand2013area,lobo2008auc}. For this reason, instead of AUC we opt to use an averaging strategy for operation thresholds \cite{wong2005s}. We compute the average accuracy for a range of operational significance levels, such as alpha from 0.20 to 0.01 for each data set and then the average obtained values for all 100 data sets for each setting. The range of alpha is obtained as follows: We vary z-score from 1.28 to 3 (equivalent to two-tailed p-value of 0.2005 to 0.0027) and then increase z-score 0.1 in each loop. 

We compare our method performance against both the STScan and the baseline approach  via control chart on ratios  described in section \ref{sec:problem}. The accuracy of methods in the identification of simulated hotspots is used as the criterion for the performance evaluation. The results are presented in Table \ref{tab:results}. As seen, EigenSpot presents a better performance in almost all settings, except low-impact hotspots. The baseline method also as expected, due to the high rate of false positives, presents the lowest accuracy. The superiority of EigenSpot over STScan possibly relies on two reasons. One reason is related to the inherent methodological difference between the EigenSpot and STScan. STScan search the whole space to find some spatiotemporal windows that the data distribution inside them has some deviation to the standard distribution models (e.g. Poisson). This strict assumption makes this approach less effective, when the data in each of sets does not exactly follow the standard distribution model or some deviation occurs by the chance. EigenSpot, instead of putting this strict restriction search for changes in the correlation patterns and therefore is less sensitive to the deviations in data distribution. The second reason could be that the EigenSpot is a shape-free method and does not search for a particular shape hotspot, while STScan looks for specific shape hotspots. Some accuracy loss in STScan relates to different shapes of the simulated hotspot. STScan looks for cylinder-shape hotspots, while the simulated hotspots are in fact cubic. 

The results also show the performance of each method against noises. We intentionally design some low-impact and size settings for evaluating the ability of the methods in handling noises and outliers. A low-size and low-impact region like impact of 1.5 and size $1 \times 1$ more seem to be an outlier or anomaly, rather than a realistic hotspot. Therefore, we expect that the methods do not detect that region as hotspot and ignore that. In other words, the detection of such hotspot shows how a method wrongly identifies the outliers and noises as hotspots. Hence, the lower accuracy in this setting reveals the better performance of the method in dealing with noises and outliers. Since the EigenSpot is a spectral method, it definitely ignores such outliers and does not report them as hotspots, while STScan is vulnerable against such circumstances. For this reason, it presents a higher performance for low-size and impact regions. In the experiment, we presumed that a hotspot with small size of $1 \times 1$ or $2 \times 2$ and low impact of 1.5 is more a noise and not a real hotspot. However, since the hotspots are simulated, this is only an assumption. We may interpret the result in other way. If we assume that impact 1.5 is not noise and reveals a real hotspot, then we can infer that STScan outperforms EigenSpot for hotspots with low impacts and sizes. 

\subsubsection{Effect of hotspot size and impact on the performance} \label{sec:evaluation}

\begin{figure}
 \begin{center}
  \includegraphics[width=.7\textwidth]{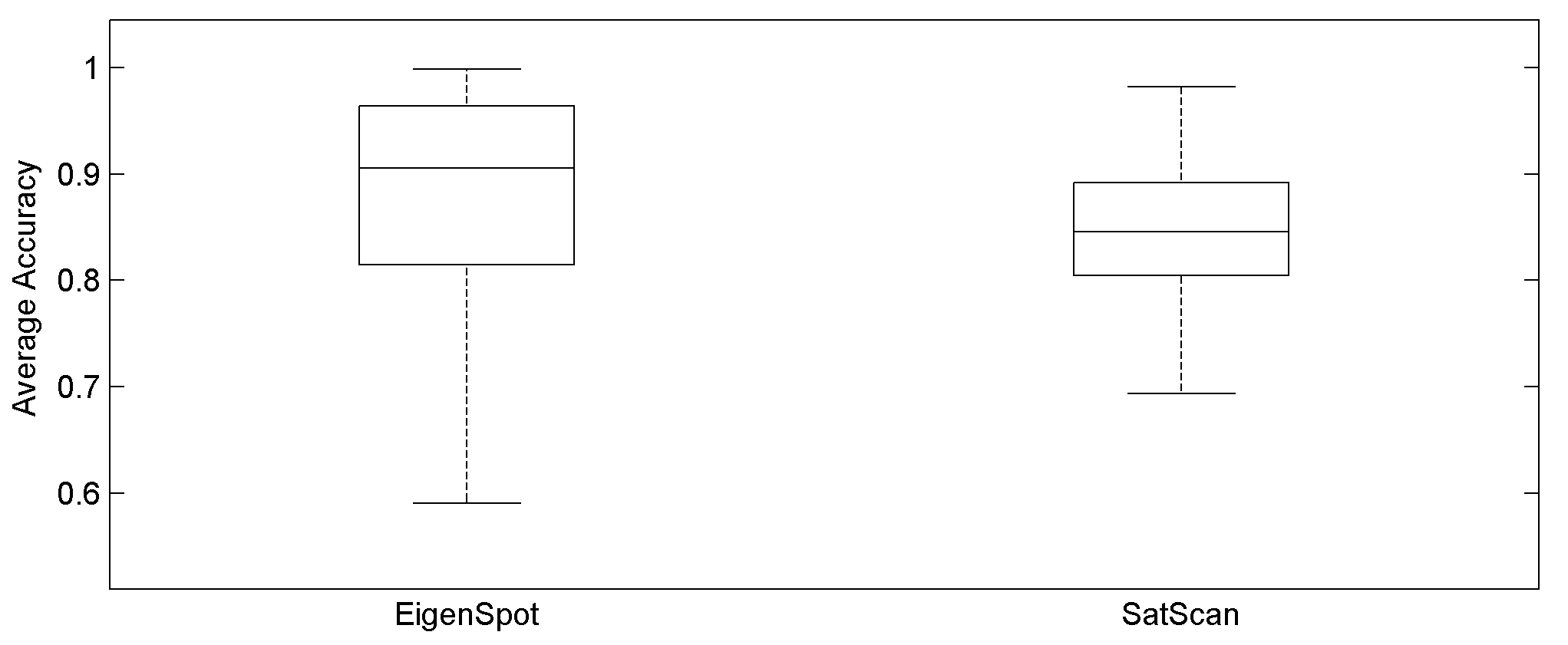}
 \end{center}
 \caption{Mean accuracy for 16000 data sets averaged for 173 $\alpha$ from 0.20 to 0.01.} \label{fig:boxplotmethods}
\end{figure}

\begin{figure}
 \begin{center}
  \includegraphics[width=\textwidth]{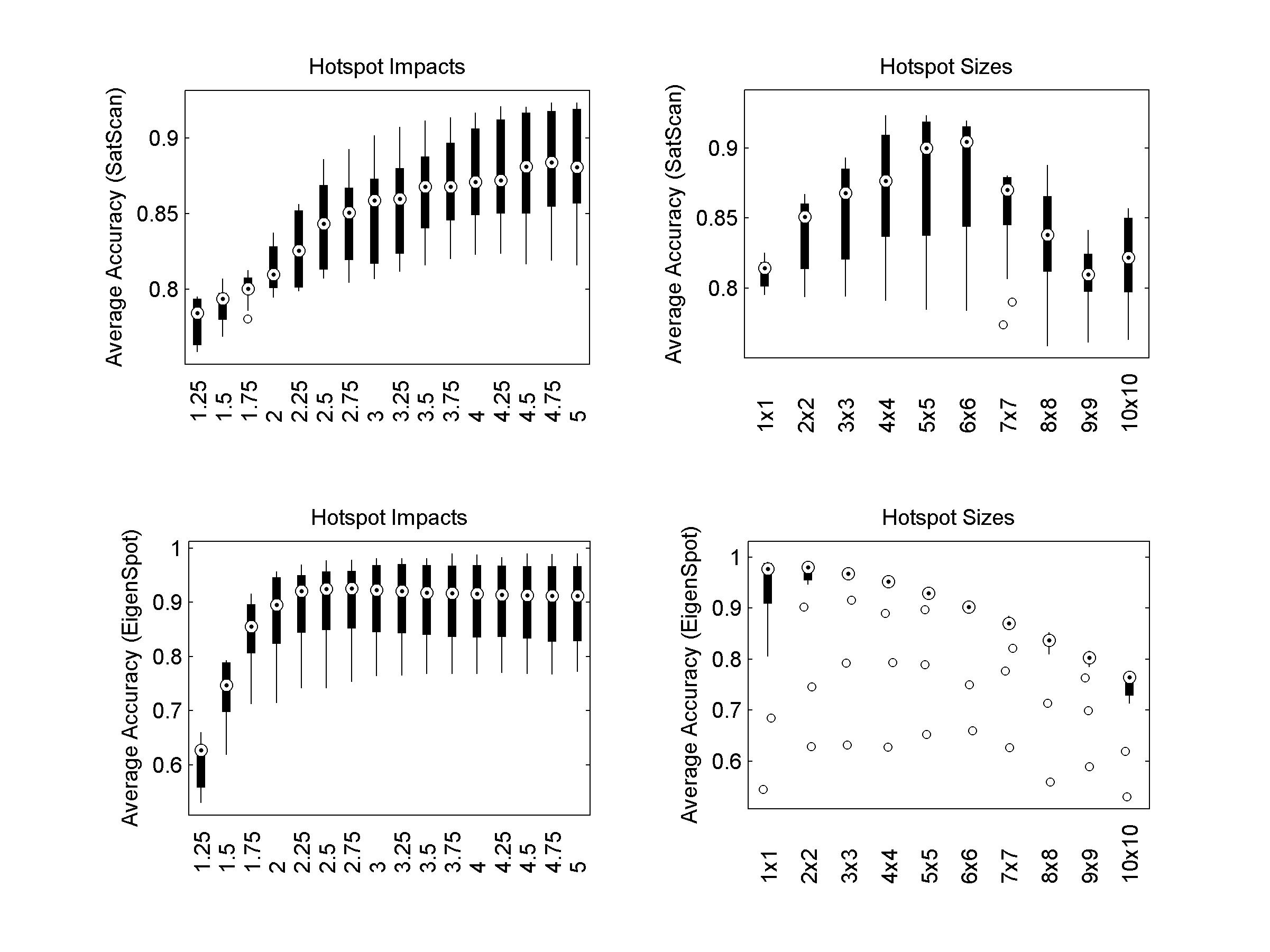}
 \end{center}
 \caption{Mean accuracy of STScan and EigenSpot corresponding to different settings for 173 $\alpha$ from 0.20 to 0.01 averaged for 100 datasets.} \label{fig:sizeimpact}
\end{figure}

\begin{table}
\begin{center}
\caption{The effect of hotspot size and impact on the performance (One-way ANOVA test).}\label{tab:anova}

  \begin{tabular}{lll}
   \hline
    Factor & STScan & EigenSpot\\ \hline
    Hotspot size  &  $p=1.6826 \times 10^{-10}$ & $p=5.9713 \times 10^{-13}$   \\ 
  	Hotspot Impact  & p=0.1834  (for impacts $\geq 2.5$) & p=0.9337 (for impacts $\geq 1.75$)  \\ 
      
  \hline
  \end{tabular}
\end{center}
\end{table}

In the previous section, we evaluated the methods performance for some limited important settings. In this section, with the same method of data generation and evaluation criterion, we study the stability of two algorithms STScan and EigenSpot for a wider range of hotspot sizes and impacts. We do not study the baseline method at this stage, since it is defeated in all previous cases. Hence, this time the hotspot size (H) varies from 1 to 10 and we test 16 different hotspot impacts (I) from 1.25 to 5 by step of 0.25. As formerly used, for each setting we generate 100 random data sets. Therefore, we generate $16000=10 \times 16 \times 100$ data sets. We then apply STScan and EigenSpot on all data sets and measure their average accuracy on 100 data sets for each setting in the operational significance levels (p-values) from 0.20 to 0.01. Figure \ref{fig:boxplotmethods} shows the result of this comparison. In order to see that whether this improvement is obtained by chance or is statistically significant, we perform a paired student's t-test between two sets of obtained performances for STScan and EigenSpot. The t-test confirms that the obtained improvement is statistically significant with $p-value=3.3591 \times 10^{-89} \approx 0$.

Figure \ref{fig:sizeimpact} shows the performance of methods against different hotspot sizes and impact. The lowest performance for EigenSpot is obtained for impacts of 1.25 and 1.5, which is more related to the noises (as was already discussed). However, we can observe that both methods relatively are robust for a hotspot impact greater than a threshold. For instance, EigenSpot is robust for impacts over 1.75 and STScan is robust for impacts over 2.5. Regarding the hotspot size, EigenSpot has a descending trend by increasing the hotspot size. This implies that by increasing the hotspot size, we should expect lower performance from EigenSpot. It makes sense, as by increasing the size of hotspot, the affected areas gradually start to seem normal and are left undetected, via a spectral method like EigenSpot. EigenSpot, however exhibits more regular behavior comparing STScan in this matter. The variance of performance is almost zero for EigenSpot during different size of hotspots, while it can vary up to 0.20 for STScan. STScan, also opposed to EigenSpot, experiences both ascending and descending trend. For hotspot sizes $1 \times 1$ to $6 \times 6$ has an ascending trend and then tend to decrease for bigger sizes. For hotspot $9 \times 9$ it has relatively the same performance as $1 \times 1$. 

To understand whether the hotspot size and impact affect the performance of the methods, we perform an ANOVA test \cite{montgomery1984design} on the obtained performance for different hotspot sizes and impacts. The null hypothesis $H_0$ is that that the mean accuracy does not change for different sizes and impacts. The test result (Table \ref{tab:anova}) confirms our initial guess that both STScan and EigenSpot become independent of hotspot impact when impact goes upper than a specific threshold. However, a very low p-values for hotspot size indicates that the performance of both EigenSpot and STScan is dependent on the hotspot size. However, as observed, both methods do not differ in their dependence on hotspot impact and size.

\subsubsection{The effect of SVD Implementation} \label{sec:evaluation}

\begin{table}
\begin{center}
\caption{Average accuracy for 1500 data sets averaged for 173 $\alpha$ from 0.20 to 0.01}\label{tab:svd}

  \begin{tabular}{lllllll}
   \hline
    Method & Compution Cost& Implementation &  Aevrage Accuracy  \\ \hline
    One-rank SVD &  $O(N^2)$ &ARPACK  & \textbf{0.8975}   \\ 
    & &IncPACK  & 0.8387 \\ \hline
  	Full SVD   &  $O(N^3)$ &LAPACK  &  \textbf{0.8429 }  \\ 
    & &PROPACK  & 0.8177 \\ 
      
  \hline
  \end{tabular}
\end{center}
\end{table}

The central technique used in EigenSpot is the SVD. Two kinds of SVD can be used for this purpose: a full-rank SVD and a low-rank SVD. Here, four of SVD implementations are chosen, two from each category and their effect is studied on the EigenSpot performance. Table \ref{tab:svd} demonstrates the average accuracy for 1500 data sets for the range of p-values from 0.20 to 0.01. As it is seen, the ARPACK implementation \cite{sorensen1997implicitly} (the default SVD implementation we use in the experiments) outperforms other methods. However, since both ARPACK and IncPACK \cite{brand2006fast} have the same computational cost, we perform an ANOVA test \cite{montgomery1984design} to see whether using IncPACK affects the performance or not. The ANOVA test shows that two sets of accuracy obtained from these two implementations are not statistically different (p-value=0.26). Therefore, we can conclude that low-rank SVD implementation used does not affect the EigenSpot performance. Concerning full-rank SVD, we observe that LAPACK \cite{anderson1999lapack} outperforms PROPACK \cite{larsen1998lanczos}, however full-rank SVD, because of its computational cost is not of our interest.

\subsection{Experiment with real data} \label{sec:real}

\begin{figure}
 \begin{center}
  \includegraphics[width=.8\textwidth]{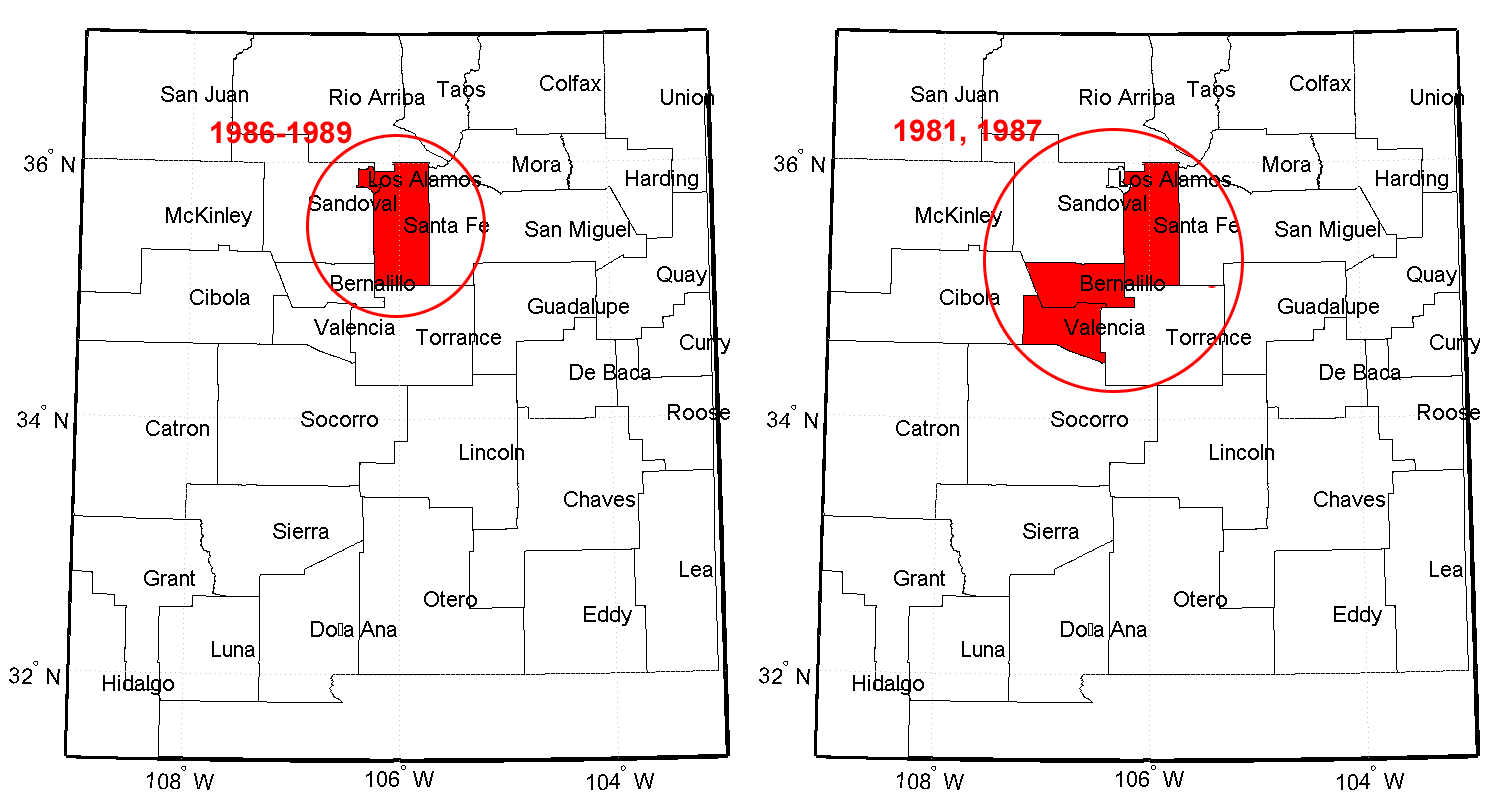}
 \end{center}
 \caption{Detected hotspot via STScan (left) and EigenSpot (Right)} \label{fig:hotspots}
\end{figure}

\begin{table}
\begin{center}
\caption{Comparison of STScan versus EigenSpot in detection of hotspots. Incidence rates were adjusted for temporal trends, age, race and sex.}\label{tab:realhotspots}

  \begin{tabular}{llll}
   \hline
    Method &  Affected Regions & Temporal Period & P-value\\ \hline
    STScan   &   Santa Fe and Los Alamos   &  1986-1989  &  0.45\\ 
    EigenSpot  &  Santa Fe, Bernalillo, Valencia  &  1981, 1987 & 0.05 \\ 
  \hline
  \end{tabular}
\end{center}
\end{table}

In this section, we study the performance of EigenSpot on a real data set. The data set which is publicly available \cite{Kulldorff2012} is provided by \textit{surveillance, epidemiology and end results (SEER) program} of the National Cancer Institute and collected by the New Mexico Tumor Registry between the years of 1973 to 1991 for 32 sub-regions of the New Mexico State, United states. There are 1175 reported cases of malignant neoplasm of the brain and the nervous system. The goal of the initial study was a response to a serious concern in 1991 in the New Mexico resident community about the correlation of wartime nuclear activities in Los Alamos with the recent brain tumor deaths in the neighborhood. The concern rapidly emerged at the local and national level and therefore became a center of attention by the local health departments. The data set was gathered, via a comprehensive review of the reported brain cancer incidence rates for the year 1973 through 1991 in order to identify the statistically significant spatiotemporal affected areas. STScan is already applied to this data \cite{kulldorff1998evaluating}. The conclusion made from the previous study shows that excess of brain cancer in Los Alamos falls within the realm of chance, which confirms the final conclusion of the New Mexico Health Department. 

In order to compare the EigenSpot with STScan, the EigenSpot is applied on the same data set. In addition to the initial study \cite{kulldorff1998evaluating} we use the adjusted incidences for temporal trends, age, race and sex. The results obtained via STScan and EigenSpot are shown in Table \ref{tab:realhotspots}. STScan reports only Santa Fe and Los Alamos in the years 1986-1989 with a relative high p-value=0.45 which indicates that there is no significant hotspot. Applying EigenSpot with $\alpha=0.05$ we could find a significant hotspot, including areas of Santa Fe, Bernalillo, Valencia as spatial components and the years 1981 and 1987 as the temporal components. However, for $\alpha=0.01$ EigenSpot does not find any hotspot. If we look the hotspot spatial positions in Figure \ref{fig:hotspots} we can find a meaningful relationship between STScan and EigenSpot results. Both candidate areas are found close the Los Alamos, the region of nuclear activities. The temporal component of 1987 also is appeared in the EigenSpot, which is located in the period detected by STScan. Based on EigenSpot result, it seems that in addition to the area close to Los Alamos and Santa Fe more areas were affected by the nuclear activities. These areas are Bernalillo and Valencia where the neighbors of Santa Fe and Los Alamos are. The very interesting point about EigenSpot is that EigenSpot opposed to STScan did not aware about the geographic relationship of the regions. STScan knows in advance that for instance whether Santa Fe and Bernalillo are neighborhood or not, while EigenSpot does not have this prior knowledge. Based on the EigenSpot result, we can infer that the effect of nuclear activities in the neighborhood has experienced two peaks during the years 1981 and 1987. It makes sense, because the initial concerns about the effect of the nuclear activities started in 1991four years right after 1987 (second detected temporal component). Indeed, EigenSpot has truly approximated the hotspot spatially and temporally very close the nuclear activity area. Most interestingly, no other meaningless hotspots are detected by EigenSpot. However, lack of strong p-value for the recognized area reveals that that the neighborhood has been under a low effect of nuclear activities in Los Alamos, but has not had an enough support to be considered alarming. If we do not consider $\alpha=0.05$ significant we can confirm initial conclusion about the random incidence of brain cancers in Los Alamos.

\section{Conclusion and future works} \label{sec:conclusion}

A new methodology for hotspot detection is proposed, which is based on two robust techniques, including matrix factorization and process control. We evaluate and compare the performance of the algorithm for detection of a single hotspot against the state-of-the art and the baseline methods through a comprehensive simulation study. The obtained results indicate a statistically significant improvement over the state-of-the-art method STScan. This improvement comes from the inherent methodological differences of the two approaches. The STScan uses the deviation in probability model as the criteria for identification of hotspots, while our approach tracks the changes the correlation patterns in spatial and temporal dimension to approximate the hotspot location. Besides, our approach is a shape-free method and contrary to STScan it is robust to the noises and outliers. 

Our approach is also much more efficient than the scan statistics-based approaches. The main benefit of our approach is that it has linear complexity, in terms of both space and time. The comprehensive comparison of scan statistics-based methods in \cite{agarwal2006spatial} reveals that any algorithms that even provide approximately optimal answers to the problem must use space linear in the input. EigenSpot provides an approximate optimal answer and is linear with space and time dimensions and it therefore, meets this requirement. 

We also study the effect of hotspot size and impact in the methods performance. Based on this result, both the STScan and EigenSpot are independent of the hotspot impact in some specific ranges. However, both methods are dependent on the hotspot size. Nevertheless, EigenSpot exhibits a more regular trend against changes in hotspot size and impact. We also study the effect of SVD implementation on the Eigenspot performance. The study shows that there is no statistical difference between two low-rank SVD implementations ARPACK and IncPACK. Therefore, SVD implementation does not affect the performance of EigenSpot. Finally, we apply EigenSpot to a real data set and compare its performance to STScan. EigenSpot, as well as the STScan recognize the affected area close to the nuclear activity area, both in space and time, however as well as STScan, cannot provide the strong statistical evidence to identify this area as hotspot. 

EigenSpot can be used as an important component in surveillance systems in particular bio-surveillance systems. Some estimations \cite{kaufmann1997economic} show that the timely detection of hotspots can save the live of thirty thousands people per day during a biogent release. It also prevents the economic cost of 250 million dollars per hour during a disease outbreak. Therefore, any early knowledge of hotspots plays an important role in improving response effectiveness. It is estimated \cite{morrison2010early} that diagnosing and controlling abnormal situations has an economic impact of at least \$10 billion annually in the United States .

Although EigenSpot is an ideal solution in terms of, both accuracy and computational cost for single hotspot detection. There is a doubt that this result is valid when multiple hotspots exist. In this work, we did not evaluate the performance of EigenSpot for multiple hotspot detection. However, theoretically we expect that STScan performs better for that purpose. Because, combining the spatial and temporal components of different hotspots together raise many false positives, which reduce the method performance, even though, this may not be considered a serious issue in disease surveillance, where in practice the most likely cluster is desired. 

There are two directions for future works. In the first direction, we intend to find a solution for adapting EigenSpot for multiple hotspot detection and in the second direction, we are going to apply EigenSpot along with visualization tools for online and real-time monitoring purposes.

\subsubsection*{Acknowledgments.} This research was supported by the Projects NORTE-07-0124-FEDER-000059/000056 which is financed by the North Portugal Regional Operational Program (ON.2 O Novo Norte), under the National Strategic Reference Framework (NSRF), through the European Regional Development Fund (ERDF), and by
national funds, through the Portuguese funding agency, Fundação para a Ciência e a Tecnologia (FCT). Authors also acknowledge the support of the European Commission through the project MAESTRA (Grant Number ICT-750 2013-612944).

\bibliographystyle{chicago}
\bibliography{ref}
\end{document}